\pdfoutput=1

\documentclass[11pt]{article}

\usepackage[]{acl}
\hypersetup{
    pdfinfo={
        Title={Meeting the Needs of Low-Resource Languages: The Value of Automatic Alignments via Pretrained Models},
        Subject={Pretrained multilingual models used for automatic word alignment.},
        Author={Abteen Ebrahimi, Arya D. McCarthy, Arturo Oncevay, Luis Chiruzzo, John E. Ortega, Gustavo A. Gimenez-Lugo, Rolando Coto-Solano, \& Katharina Kann},
    }
}

\usepackage{times}
\usepackage{latexsym}

\usepackage[T1]{fontenc}

\usepackage[utf8]{inputenc}

\usepackage[shrink=5]{microtype}  

\usepackage{inconsolata}

\usepackage{adjustbox}
\usepackage{booktabs}
\usepackage{multirow}
\usepackage{svg}
\usepackage{subcaption}
\usepackage{caption}
\usepackage[nameinlink,capitalize,noabbrev]{cleveref}

\crefname{equation}{equation}{equations}   
\crefname{footnote}{footnote}{footnotes}   
\crefname{section}{\S}{\S\S}
\Crefname{section}{\S}{\S\S}    
\crefformat{section}{#2\S#1#3}  
\Crefformat{section}{#2\S#1#3}
\crefrangeformat{section}{\S\S#3#1#4--#5#2#6}
\Crefrangeformat{section}{\S\S#3#1#4--#5#2#6}
\crefformat{section}{#2\S#1#3}  
\crefrangeformat{section}{\S\S#3#1#4--#5#2#6}

\newcommand\blfootnote[1]{
  \begingroup
  \renewcommand\thefootnote{}\footnote{#1}%
  \addtocounter{footnote}{-1}%
  \endgroup
}

\newcommand{\lang}[1]{\textsc{#1}}

\title{Meeting the Needs of Low-Resource Languages:\\ The Value of Automatic Alignments via Pretrained Models}

\author{Abteen Ebrahimi${ }^{\diamondsuit}$ \quad
Arya D. McCarthy${ }^{\nabla}$ \quad
Arturo Oncevay${ }^{\heartsuit}$ \\
\textbf{Luis Chiruzzo${ }^{\triangle}$ \enskip
John E. Ortega${ }^{\Omega}$ \enskip
Gustavo A. Giménez-Lugo${ }^{\clubsuit}$} \\
\textbf{Rolando Coto-Solano${ }^{\phi}$ \enskip
Katharina Kann${ }^{\diamondsuit}$} \\
${ }^{\diamondsuit}$University of Colorado Boulder \quad
${ }^{\nabla}$Johns Hopkins University \quad
${ }^{\heartsuit}$University of Edinburgh \\
${ }^{\triangle}$Universidad de la República, Uruguay \quad
${ }^{\Omega}$Northeastern University \\
${ }^{\clubsuit}$Universidade Tecnológica Federal do Paraná \quad
${ }^{\phi}$Dartmouth College \quad
}

\begin{document}
\maketitle
\begin{abstract}
 Large multilingual models have inspired a new class of word alignment methods, which work well for the model's pretraining languages. However, the languages most in need of automatic alignment are low-resource and, thus, not typically included in the pretraining data. In this work, we ask: \textit{How do modern aligners perform on unseen languages, and are they better than traditional methods?} We contribute gold-standard alignments for Bribri--Spanish, Guarani--Spanish, Quechua--Spanish, and Shipibo-Konibo--Spanish. With these, we evaluate state-of-the-art aligners with and without model adaptation to the target language. Finally, we also evaluate the resulting alignments extrinsically through two downstream tasks: named entity recognition and part-of-speech tagging. We find that although transformer-based methods generally outperform traditional models, 
 the two classes of approach remain competitive with each other.\looseness=-1
\end{abstract}

\section{Introduction}
\begin{figure}
    \centering
    \includegraphics[width=\linewidth]{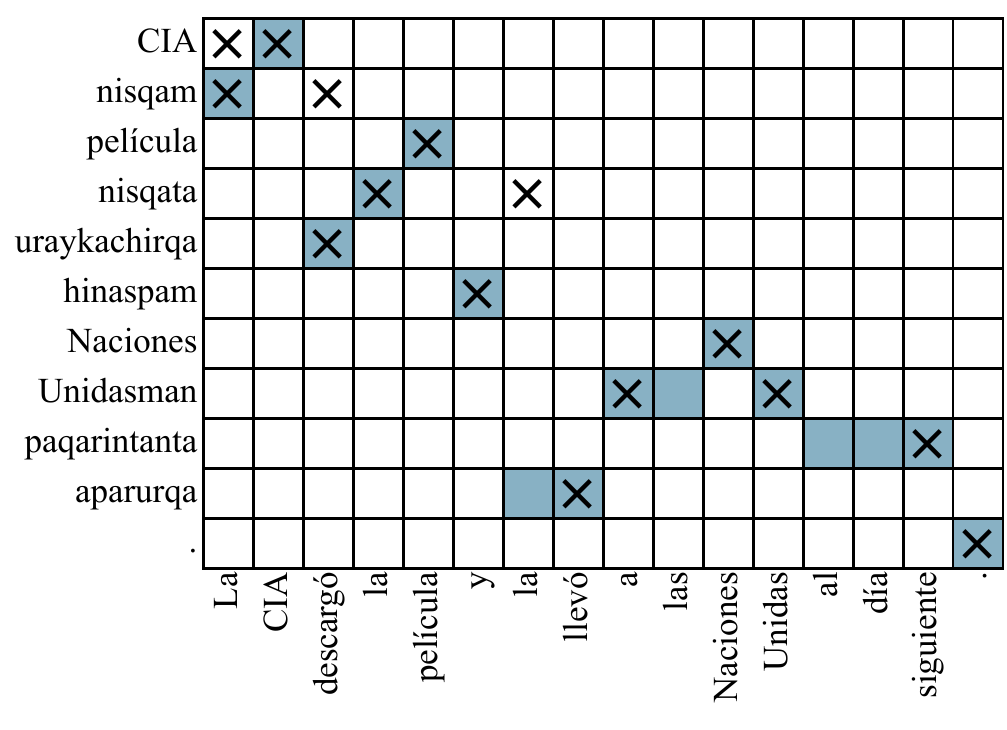}
    \caption{A word alignment between Quechua and Spanish (shaded), as well as mBERT+TLM's predicted alignment (marked by \(\times\)'s). FastAlign and Giza++ cannot take advantage of surface features of proper names and borrowings. We evaluate alignments intrinsically via AER and extrinsically with POS-tagging and NER models learned on annotations projected across alignments from Spanish.}
    \label{fig:alignment}
\end{figure}

Word alignment is a valuable tool for extending the coverage of natural language processing (NLP) applications to low-resource languages through, e.g., statistical machine translation \citep[SMT; ][]{koehn-knowles-2017-six,duh-etal-2020-benchmarking} or annotation projection \cite{yarowsky-etal-2001-inducing,smith-smith-2004-bilingual,nicolai-etal-2020-fine,eskander-etal-2020-unsupervised}. The traditional approach for generating alignments has been with statistical methods such as Giza++ \cite{Och2003ASC} and FastAlign \cite{Dyer2013ASF}, which provide strong alignment quality while remaining quick and lightweight to run. Recently, new methods have been proposed which extract alignments from massive \textit{pretrained multilingual models}, and outperform these longstanding methods \cite{dou2021word}. \blfootnote{Our code and data can be found at \url{https://github.com/abteen/alignment}.}

However, results on other NLP tasks, such as part-of-speech (POS) tagging and named-entity recognition (NER), have shown that, while pretrained models generally work well out-of-the-box for high-resource languages, performance is far lower for low-resource ones, particularly those which are unseen during pretraining \cite{pires-etal-2019-multilingual, wu-dredze-2020-languages,muller-etal-2021-unseen,lee-etal-2022-pre}. Models can be adapted \cite{gururangan-etal-2020-dont, chau-etal-2020-parsing} to improve performance, but this comes with a large computational cost. 
Given these two considerations, for \emph{unseen} low resource languages it remains unclear (1) whether modern neural approaches based on adapted pretrained models generate higher-quality alignments than traditional approaches and (2) if so, whether the quality difference is severe enough to justify the additional computational cost.

We investigate this by collecting gold-standard alignments for Bribri, Guarani, Quechua, and Shipibo-Konibo. These languages are low-resource and unrepresented in the pretraining data of popular models---a relevant real-world scenario. In addition to intrinsically evaluating alignment quality, we measure the downstream utility of each method for training POS-tagging and NER models by annotation projection.

We find traditional and neural methods to be competitive, but pretrained models result in slightly lower alignment error rates and stronger downstream task performance, even for initially unseen languages. Through further analysis, we also find that adaptation may be a more reliable approach given minimally available resources. Taken together, these results indicate that alignment from multilingual models can indeed be a valuable tool for low-resource languages, but traditional approaches continue to be a strong option and should still be considered for practical applications.

\section{Related Work}

\paragraph{Alignment}
Word alignment is a long studied task, with origins in the IBM models for statistical machine translation \cite{brown-etal-1993-mathematics}, which are the basis of Giza++ \cite{Och2003ASC} and FastAlign \cite{Dyer2013ASF}. As these approaches can only generate one-to-many alignments, models are trained in both forward and reverse directions (reversing the role of source and target), and final alignments are created via symmetrization heuristics \cite{Och2000ImprovedSA, koehn-etal-2005-edinburgh}; other approaches explicitly symmetrize during training \citep{matusov-etal-2004-symmetric,liang-berkeley}.\footnote{The poor estimation of rare words' translation parameters also motivates symmetrization; without this, rarely observed words become \emph{garbage collector words} \citep{moore-2004-improving}.}
While these models rely on only position and word identity information, subword information can be integrated without requiring costly inference \citep{berg-kirkpatrick-etal-2010-painless}, leading to better parameter estimation for rare words.
Alignments can also be extracted from neural translation models \cite{chen-etal-2020-accurate, zenkel-etal-2020-end}. 

Word alignment also enables annotation projection \cite{yarowsky-ngai-2001-inducing,yarowsky-etal-2001-inducing} which can offer strong performance, particularly for low-resource languages \cite{buys-botha-2016-cross, ortega2018using, nicolai-yarowsky-2019-learning,nicolai-etal-2020-fine,eskander-etal-2020-unsupervised}.

\paragraph{Multilingual Transformer Models} Pretrained multilingual models \cite{devlin-etal-2019-bert, Lample2019CrosslingualLM, conneau-etal-2020-unsupervised,xue-etal-2021-mt5} have become the de facto standard approach for cross-lingual transfer. In general, these models are an extension of their monolingual variants, created by including data from many languages in their pretraining. They rely on a subword vocabulary \cite{kudo-richardson-2018-sentencepiece} which jointly spans all of the pretraining languages.
Models are pretrained using a masked language modeling (MLM) objective and a translation language modeling \citep[TLM;][]{Lample2019CrosslingualLM} objective that uses parallel sentences.
Outside of continued pretraining \cite{gururangan-etal-2020-dont}, models can be adapted using Adapters \cite{pfeiffer-etal-2020-mad} or through vocabulary adaptation \cite{wang-etal-2020-extending, hong-etal-2021-avocado}. Word alignment methods which depend on these models have also been proposed \cite{jalili-sabet-etal-2020-simalign,nagata-etal-2020-supervised}; we focus  on AWESoME align \cite{dou2021word} because it outperforms other unsupervised methods.

\section{Experiment 1: Intrinsic Evaluation}

\subsection{Experimental Setup}
\paragraph{Languages} We focus on four Indigenous languages spoken in the Americas for our experiments. \textbf{Bribri (bzd)} is a tonal language in the Chibchan family spoken by approximately 7000 people in Costa Rica. \textbf{Guarani (gn)} is a polysynthetic language in the Tupi--Guarani family spoken by around 6 million people across South America. \textbf{Quechua (quy)} is a family of Indigenous languages---from which we study Quechua Chanka---spoken across the Peruvian Andes by over 6 million people, and \textbf{Shipibo-Konibo (shp)} is a language spoken by around 30,000 people in Peru, Bolivia, and Brazil \cite{cardenas-zeman-2018-morphological}. The latter three languages are agglutinative. 

\paragraph{Training Data} For training, we use the parallel data between Spanish and our languages described by \citet{mager-etal-2021-findings}.\footnote{Although parallel, digitized Bibles exist for over 1600 languages \citep{mccarthy-etal-2020-johns}, groups may object to annotating the Bible for historical or cultural reasons.} We note that there is a distinct difference in the amount of unlabeled data available within the four languages: Guarani and Quechua have considerably more data available. These two languages also have monolingual text available in Wikipedia, which we extract using WikiExtractor \cite{Wikiextractor2015}. 
The exact number of parallel and monolingual sentences for all languages is shown in \cref{tab:parallel_data_details}. 
    
\paragraph{Evaluation Data} To create gold standard alignments for evaluation, we sample multi-way parallel examples from AmericasNLI \cite{ebrahimi-etal-2022-americasnli},
allowing for multi-parallel alignments \citep{xia-yarowsky-2017-deriving} across all languages. Samples for the development and test sets are taken from their respective splits in the AmericasNLI dataset. Development examples were collected first, manually checked, and corrected. Examples with misalignments in punctuation, numbers, or named entities were not used. After a period of development with this data, the test set of 50 examples was collected and manually verified. Annotations were collected using JHU's open-source Turkle platform.\footnote{\url{https://github.com/hltcoe/turkle}} We ask annotators to only mark \textit{sure} alignments. Additional discussion on data collection and the test set can be found in \cref{ethics}.

\paragraph{Metrics} We evaluate automatic alignments via alignment error rate \citep[AER; ][]{Och2000ImprovedSA}. Because we only collect sure alignments, this is equivalent to the balanced F-measure \cite{fraser-marcu-2007-squibs}. We give additional metrics in \cref{tab:prf_test}.

\subsection{Models}
\paragraph{Traditional Aligners} We use 
Giza++ \cite{Och2003ASC} and FastAlign \cite{Dyer2013ASF} as our traditional aligners. Giza++ is based on IBM Models 1--5 \cite{brown-etal-1993-mathematics}. FastAlign \cite{Dyer2013ASF} is a re-parameterization of  IBM Model 2. We use the implementation and hyperparameters of \citet{zenkel-etal-2020-end}, which relies on MGiza++ \cite{Gao2008ParallelIO} and the standard FastAlign package. Both approaches run on CPUs, and their training time ranges between 6 seconds to 3 minutes for FastAlign, and 43 seconds to 22 minutes for Giza++. We use the union of the forward and reverse alignments, as this symmetrization heuristic offers the best result for all languages on the development set. We show the performance of other heuristics in \cref{tab:growing_heuristics}.

\paragraph{Neural Aligners}

AWESoME \cite{dou2021word} identifies alignment links by considering cosine similarities between hidden layer representations of tokens in a neural encoder. We consider two such encoders: mBERT \cite{devlin-etal-2019-bert} and XLM-R \cite{Liu2019RoBERTaAR}, and we use the default AWESoME configuration to extract alignments. We give layer-by-layer alignment performance in \cref{fig:heatmap}.

\paragraph{Model Adaptation} We experiment with three adaptation schemes based on continued pretraining (+TLM, +MLM-T, and +MLM-ST) which rely on unlabeled data and further train the model using MLM \cite{gururangan-etal-2020-dont} before alignments are extracted. We focus on these objectives as they have been used by prior work for general model adaptation, and they work well in situations with limited resources \cite{ebrahimi-kann-2021-adapt}. As we have access to bitext between Spanish and the target languages, for the +TLM scheme each example is the concatenation of a Spanish sentence with its translation. 
For +MLM-\textbf{T} we adapt using solely the \textbf{t}arget side of the available data, and for +MLM-\textbf{ST} we adapt on both the \textbf{s}ource and \textbf{t}arget; however, this data is treated as monolingual data and not explicitly aligned. 
+MLM-WT denotes target language adaptation which includes Wikipedia data. 
The duration of adaptation depends on the GPU and method used; it ranges from around 6 minutes for Bribri to 4 hours for Quechua. 
We provide additional training details in \cref{training_info}. 
\begin{table}[t]
\centering
\small
\setlength{\tabcolsep}{1.8pt}
    \begin{tabular}{@{} l l cccc @{\quad} c @{}}
    \toprule
    Model & Method & \lang{bzd} & \lang{gn} & \lang{quy} & \lang{shp} & \lang{avg.} \\
    \midrule
    AWESoME  & BL  & 70.03 & 63.13 & 67.02  & 60.41 & 65.15  \\
    (mBERT)         & +MLM-T & 68.95 & 49.68 & 46.59  & 58.17 & 55.85  \\
                   & +MLM-ST & 70.63 & 50.25 & 42.52  & 58.66 & 55.52  \\
                   & +TLM   & \underline{58.43} & \textbf{43.10} & \textbf{36.96}  & \textbf{52.34} & \textbf{47.71}  \\
    \addlinespace 
    AWESoME  & BL  & 80.15 & 73.11 & 75.24 & 69.21 & 74.43 \\
    (XLM-R)         & +MLM-T & 76.89 & 65.44 & 53.65 & 65.16 & 65.29 \\
                   & +MLM-ST& 77.53 & 64.55 & 52.90 & 66.56 & 65.39 \\
                   & +TLM   & \underline{74.90} & \underline{58.84} & \underline{43.25} & \underline{63.48} & \underline{60.12} \\
    \addlinespace
    FastAlign  & Union &\textbf{51.40} & \underline{43.52} & \underline{54.06} & \underline{54.67} & \underline{50.91} \\
    Giza++ & Union & 55.61 & 49.92 & 66.01 & 60.84 & 58.10 \\
    \midrule
    mBERT  & +MLM-WT  &-& \textit{\textbf{40.00}} & 46.00 & - & 43.00\\
    XLM-R  & +MLM-WT  &-& 52.27 & 48.83 &- & 50.55\\
    \bottomrule
    \end{tabular}
    \caption{AER, in percentages, for each language and method. The best overall result for each language is bolded, while the best model within each method is underlined. We separate results which use Wikipedia, as they are not directly comparable.}
\label{tab:main_aer_result}

\end{table}

\subsection{Results} 

\paragraph{Traditional vs.\ Neural Aligners} We present results 
in \cref{tab:main_aer_result}. 
The best traditional method is FastAlign, and the best neural approach is with mBERT+TLM. Comparing the two, we see that the lowest error rate is achieved with the neural approach for all languages except for Bribri, where FastAlign offers 7.03\% absolute improvement. Of the other three languages, the performance for two is close: the difference in performance for Guarani is only 0.42\% and 2.33\% for Shipibo-Konibo. For Quechua, +TLM improves over FastAlign by 17.10\%.

\paragraph{Comparing Adaptation Strategies} 
With mBERT, +MLM-T improves performance over the non-adapted baseline by 9.30\% on average, with +MLM-ST increasing this gain to 9.63\% and +TLM offering the highest improvement of 17.44\%, consistent with prior work on \textit{seen} languages \cite{dou2021word}. Per language, the largest and smallest gains are for Quechua (30.06\%) and for Shipibo-Konibo (8.07\%); intuitively, gains from adaptation are proportional to the size of the adaptation data.  
For XLM-R, we again see relative gains from adaptation, with +TLM offering the highest performance increase.

\paragraph{Additional Monolingual Data} Neural approaches can easily benefit from additional monolingual data. Adding Wikipedia data results in the highest performance for Guarani, outperforming the previous best approach by 3.1\%. In contrast, while the additional data for Quechua does help relative to +MLM-T, it does not outperform +TLM. This difference in performance may be due to the relative sizes of the additional data; the Guarani Wikipedia has 1.3\(\times\) as many tokens as the target-side parallel data, while the Quechua Wikipedia only has 0.5\(\times\) as many.

\begin{table}[t]
\centering
\small
\setlength{\tabcolsep}{8.5pt}
    \begin{tabular}{@{} l l r r @{}}
    \toprule
    Model & Train Source & \lang{POS} & \lang{NER} \\
    \midrule
    mBERT  & en  & 10.36 & 46.64  \\
            & es & 19.82 & 49.18   \\
            \midrule
                +TLM & es & 36.94 & 49.62 \\
    +MLM-T & es & 34.69 & \textbf{55.25} \\
    +MLM-ST & es & 33.78 & 52.34 \\
    \midrule
    mBERT        & mBERT & 31.53 & 47.54 \\
            & +MLM-T & 38.29 & 47.97 \\
            & +MLM-ST & \textbf{42.34} & 49.80 \\
            & +TLM &40.99 & 49.80 \\
            & FastAlign & 37.84& 46.55 \\
            & Giza++ &39.19&48.33 \\

    \bottomrule
    \end{tabular}
\caption{POS tagging (accuracy) and NER results (F1) for Guarani. \textit{Model} denotes if baseline or adapted mBERT is used. \textit{Train Source} defines the training data used for finetuning; language codes indicate training on original data, while alignment methods denote how a projected training set was created.}
\label{tab:downstream_results}
\end{table}
\begin{figure*}[ht]
    \centering
    \begin{subfigure}[b]{0.49\textwidth}
    \centering
    \includegraphics[width=\textwidth]{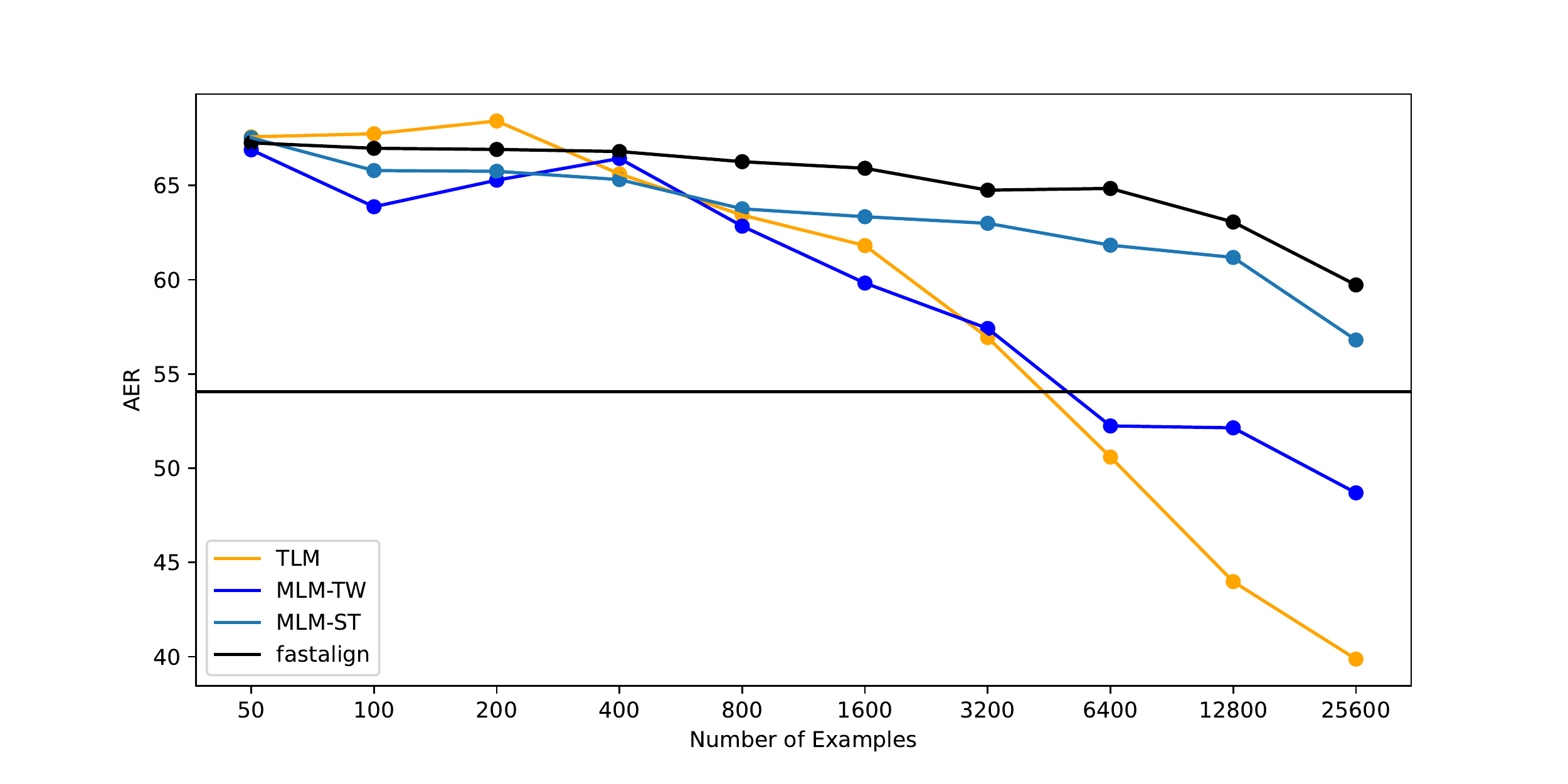}
    \caption{Subset Analysis}
    \label{fig:subset_analysis_subfig}
    \end{subfigure}
    \begin{subfigure}[b]{0.49\textwidth}
    \centering
    \includegraphics[width=\textwidth]{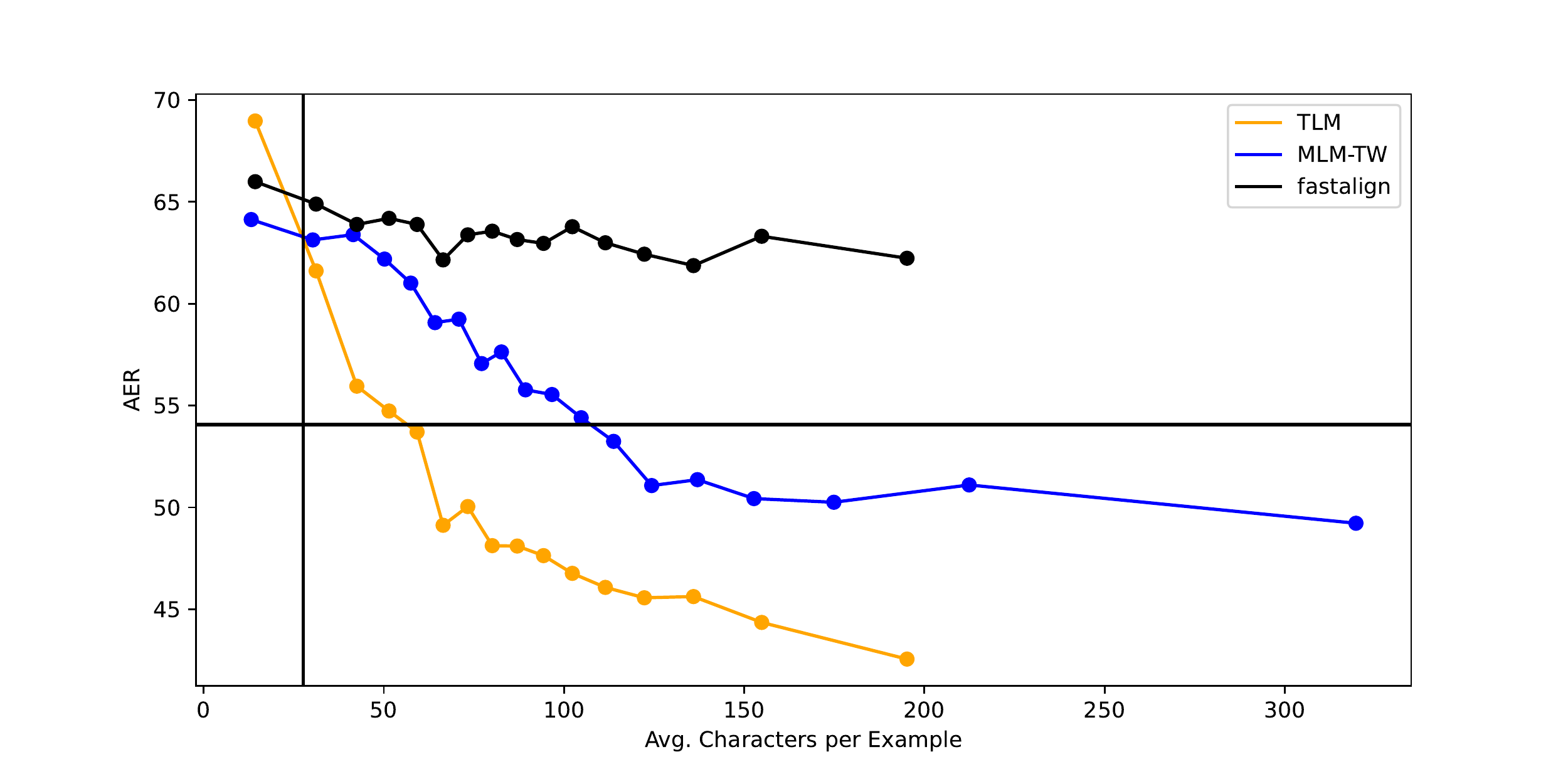}
        \caption{Length Analysis}
        \label{fig:length-analysis-labels}
    \end{subfigure}

    \caption{Plots for data analysis. In \cref{fig:length-analysis-labels}, a vertical line denotes the average example length for Bribri.}
    \label{fig:analysis_plots}
\end{figure*}
\section{Experiment 2: Extrinsic Evaluation}
We further compare aligner performance extrinsically by evaluating downstream task performance when using a projected training set. We consider two tasks: NER and POS tagging. 

\subsection{Experimental Setup}

\paragraph{Data} Due to the limited availability and quality of evaluation datasets, we focus on Guarani for this experiment. We use the test set provided by \citet{rahimi-etal-2019-massively} for NER and Universal Dependencies \cite{nivre-etal-2020-universal} for POS. For experiments where we finetune directly on English or Spanish, we use the provided training data. 

\paragraph{Annotation Projection} To create the projected training sets, we first annotate the (unlabeled) Spanish parallel data with Stanza \cite{qi2020stanza} and generate bidirectional alignments using each method. We then project the tags from Spanish to Guarani using type and token constraints as described by \citet{buys-botha-2016-cross}.

\paragraph{Models} For baseline performance, we finetune mBERT on the provided English and Spanish training sets for each task. Additionally, we also finetune adapted versions of mBERT on Spanish training data -- English is omitted as performance is worse and adaptation data is in Spanish. Finally, we evaluate performance when finetuning mBERT on the training sets created through projection. 

\subsection{Results}
We present results for both tasks in \cref{tab:downstream_results}. 
\paragraph{POS} For POS tagging, the baseline zero-shot performance is extremely poor, and we see a minimum increase of 11.71\% accuracy when using any projection method. Giza++ outperforms FastAlign, as well as projection with +MLM-T, however the best performance is achieved with +MLM-ST, with +TLM offering the second best result. While the ordering of methods changes, the best performance is still achieved with the neural approaches, consistent with the results of Experiment 1.

\paragraph{NER} For NER, baseline performance is high: inspecting the data shows that many entities have English or Spanish names, and as multilingual models already have knowledge of these two languages, standard aligners with projection may not effectively leverage surface word-form clues. However, they remain a valuable indication of alignment quality. Among the projection-based approaches, we find that using Giza++ again outperforms +MLM-T and FastAlign but falls short of +MLM-ST and +TLM.

Overall, considering what both downstream tasks indicate regarding alignment quality, 
neural models adapted using Spanish and target-language data---either sentence-aligned or unaligned---consistently outperform traditional methods.

\section{Analysis} As data for low-resource languages often varies considerably in both amount and length, we consider two additional analysis experiments which control for these factors.  
We focus solely on Quechua, as it has the most parallel data available. Results are presented in \cref{fig:analysis_plots} with numerical results in \cref{tab:subset_analysis_results,tab:length_analysis_results}.

\paragraph{Subset Analysis} For this analysis, we ask how the performance of neural alignment depends on the amount of data and with how much data it surpasses traditional approaches. 
We subsample the adaptation data, and use this to extract alignments using both FastAlign and AWESoME. Results for this experiment can be seen in \cref{fig:subset_analysis_subfig}. For reference, we also plot the AER obtained when using FastAlign on all the available training data as an upper bound for the performance of the traditional approaches. In the smallest extreme, all methods are roughly equivalent. However, as the number of examples increases, adaptation using +TLM and +MLM-WT improves at a faster rate than other approaches: with only 6400 sentence pairs, these approaches overtake the best expected performance of FastAlign.

\paragraph{Length Analysis} Aligner performance may not only be affected by the total number of examples available, but also by the length of these examples. This is doubly relevant for low-resource languages, as resources may be limited to sources which do not contain long (or even complete) sentences. To see how the performance of each method may vary when faced with examples of different lengths, we sort the unlabeled data by the number of characters, and partition the examples in groups of 7508, the total number of examples available for Bribri. We choose this amount as it is representative how much data may be available for other low-resource languages. As before, the expected upper bound FastAlign performance is denoted. 
For the shortest group, all methods are similar; however, AWESoME alignments improve with longer sequences, with +TLM showing the quickest decrease in error rate. We attribute the improved AER when adapting using longer sequences to the increased number of tokens available for adaptation. For Quechua, the performance of AWESoME align is sensitive to both the number of examples and sequence length. In contrast, FastAlign only shows a small improvement as example length increases.

\section{Conclusion}
In this work, we have investigated the performance of modern word aligners versus classical approaches for 
languages \textit{unseen} to pretrained models. While classical methods remain competitive, the lowest AER on average is achieved by modern neural approaches. However, using these models comes with a larger computational cost. Therefore, 
the trade-off between training requirements and overall performance must be considered. 
If access to computing resources is limited or training time is a factor, classical approaches remain a viable approach which should not be discounted.

\section*{Ethics and Limitations}
\label{ethics}

\subsection*{Ethics Statement}

When collecting data in an Indigenous language, it becomes vital that the process does not exploit any member of the community or commodify the language \cite{schwartz-2022-primum}. Further, it is important that members of the community benefit from the dataset. While the creation of a word alignment dataset will not directly impact community members, we believe that it can contribute to the development of tools, such as translation systems, that can be directly beneficial, and that increasing the visibility of these languages within the research community will further spur the creation of useful systems. Our annotations were created by either co-authors of the paper or by native speakers of the languages, who were compensated at a rate chosen with the minimum hourly salary in their respective countries taken into account. 

\subsection*{Limitations}
\paragraph{Test Set Size}\label{test_set_size} One limitation of our work is the size of the evaluation set used for our main results. This arises from the general difficulty in collecting annotations and data for low-resource, and particularly Indigenous languages. The size of the test set was chosen to balance the trade-off between the cost of annotation collection and experimental validity. Fortunately, for the task of word alignment the main metric used to summarize performance---alignment error rate---does not depend directly on the number of examples in the evaluation set, but the total number of alignments, of which there are a sufficiently high number in our evaluation set. However, even when only considering the number of examples, our test set is still within the same order of magnitude as other widely used word alignment evaluation sets, such as the Romanian--English test set which consists of 248 examples \cite{mihalcea-pedersen-2003-evaluation}, and the English--Inuktitut and English--Hindi test sets which have 75 and 90 examples each, respectively \cite{martin-etal-2005-word}. 

We run a small experiment to gain insight into how much precision is lost when using a test set of size 50, versus 248, which we choose as this is the size of the widely used Romanian--English test set mentioned above. We take 100 independent samples without replacement from the Romanian-English test set, each of size 50, and evaluate the performance of FastAlign and AWESoME align. For FastAlign, we use the training data defined by \citet{mihalcea-pedersen-2003-evaluation}, and for Awesome, we use mBERT with no additional finetuning. The distributions of AER are shown in \cref{fig:violin}, with summary statistics in \cref{tab:subset_summary_stats}. We can see that the standard deviation of both distributions is relatively low, around 2\%. At the extremes, we see a difference of \(-\)4.70\% and \(+\)4.90\%, and \(-\)4.28\% and \(+\)6.4\% for FastAlign and AWESoME align respectively, between the min/max values of our distribution as compared to the whole set AER. Considering these points, we believe that the size of our evaluation set does not invalidate our experimental results and main conclusions; however, we note that additional care must be taken when comparing specific models whose performances are close together, particularly when this performance is low or close to random. 

\paragraph{Test Set Domain} Other limitations of our work arise from the sources of data used. Annotations were done using sentences sampled from AmericasNLI, which itself is a translation of XNLI. As such, any errors from the original XNLI dataset, which may have propagated through translation, will persist in our dataset as well (annotators were given the option to modify target language sentences to correct any errors). Furthermore, due to translation, the sentences may not be directly representative of a natural utterance which would be spoken by members of the communities.

\paragraph{Language Selection} The languages we highlight in this work are true low-resource languages, and present challenges commonly faced by other low-resource languages. Namely, these languages have a relatively small amount of easily available and clean unlabeled data, are typically unseen from most released pretrained models, and are morphologically different from typically used source languages. However, one feature of these languages which may inflate aligner performance is the language script: all of our target languages share the same script with the two source languages which we use. This may lead to higher occurrences of shared words or entities, making alignment easier. As such, our results may not generalize fully to other low-resource languages which have a different script from the source languages, or which may have a script which is unseen to the underlying pretrained model. 

\section*{Acknowledgements}
We would like to thank Roque Helmer Luna-Montoya (Academia Mayor de la Lengua Quechua in Cuzco, Perú) and Richard Castro Mamani (Universidad Nacional de San Antonio Abad and Hinantin Software) for their help in annotating and verifying the Quechua--Spanish alignments. We would also like to thank Liz Karen Chavez Sanchez for annotating the Shipibo-Konibo--Spanish alignments. A.D.M. is supported by an Amazon Fellowship and a Frederick Jelinek Fellowship.

\bibliography{anthology,custom}
\bibliographystyle{acl_natbib}
\clearpage
\appendix
\counterwithin{figure}{section}  
\counterwithin{table}{section}  
\onecolumn

\section{Training Details and Hyperparameters}
\label{training_info}
We compare two data loading strategies for adaptation: a naïve approach where each example in the dataset represents an example in the loaded training examples, and a packing strategy following the \texttt{FULL-SENTENCES} approach of \citet{Liu2019RoBERTaAR}. We use the hyperparameters described by \citet{ebrahimi-etal-2022-americasnli} -- a learning rate of 2e-5, batch size of 32, and warmup ratio of 1\% -- however due to the different loading strategy we tune the total amount of training time. We experiment with 40 and 80 epochs of training, using the alignment development set to select the final hyperparameters. For both MLM-T and MLM-ST we find that packing sequences yields better results, however for +TLM we use the naïve strategy to preserve sentence alignment. We use packing by default for Wikipedia data, due to the length of extracted documents. For all adaptation methods we find that training for 80 epochs is best, except for +MLM-ST, which we train for 40. We train with 1 Nvidia A100 or 2 V100 GPUs. Due to the computational cost associated with pretraining, we only conduct one model run for each language and method. We pretrain our models using Huggingface \cite{wolf-etal-2020-transformers}.

\paragraph{Training Time} As mentioned in Section 3.2, for adaptation the training duration depends on the GPU and method used, with times ranging from around 6 minutes for Bribri to 4 hours for Quechua. For the statistical approaches, both run solely on CPUs, and their training time ranges between 6 seconds to 3 minutes for FastAlign, and 43 seconds to 22 minutes for Giza++. However, GPU availability is not always certain -- to roughly compare training times given a more restricted setting, we run our adaptation experiments without access to any GPUs, and compute an estimate for the total training time using only CPUs as approximately 2 weeks. 

\begin{table}[h]
    \begin{adjustbox}{width=\textwidth}
    \centering
    \begin{tabular}{l | c| c c c c c c c }
    \toprule
    & Whole Set AER & Avg. AER & AER Std. & Min AER & 25\% & 50\% & 75\% & Max AER \\
    FastAlign & 35.00 & 35.09 & 1.94 & 30.30 & 33.70 & 35.00 & 36.23 & 39.90 \\
    Awesome & 28.23 & 28.26 & 2.04 & 23.95 & 26.66 & 28.12 & 29.71 & 34.63 \\
    \bottomrule
    \end{tabular}
    \end{adjustbox}
    \captionof{table}{Summary statistics for subsample AER distribution.}
    \label{tab:subset_summary_stats}
    
\end{table}

\begin{figure*}[h]
\centering
\includegraphics[width=0.75\textwidth]{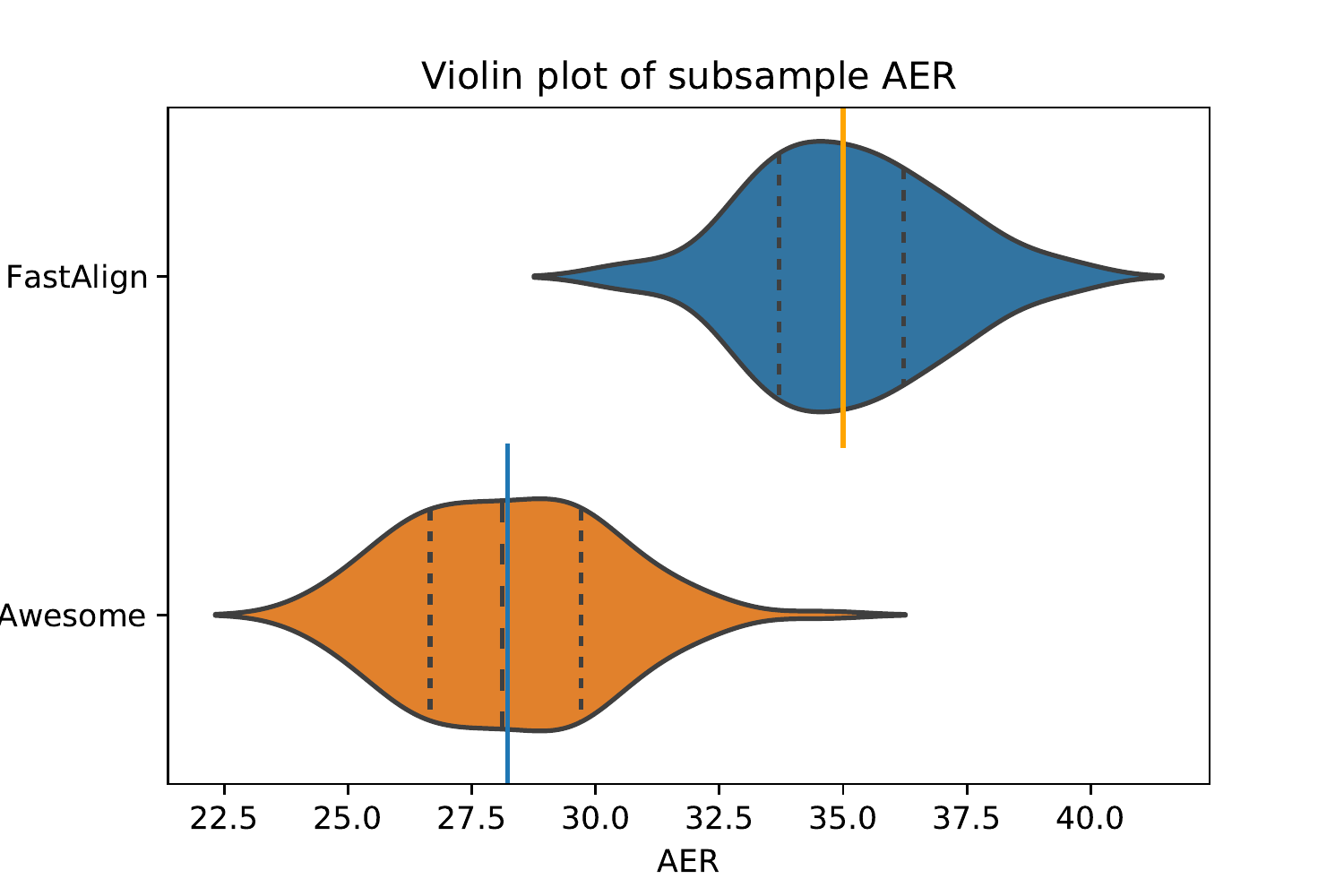}
\caption{Distribution of AER when using FastAlign and AWESoME align to evaluate subsets of size 50 taken from a complete evaluation set of size 248. Quartiles are displayed using dashed lines, while inverted colors represent the AER calculated when evaluating on the complete set.}
\label{fig:violin}
\end{figure*}
\clearpage
\section{Dataset Features}

\begin{table}[h]
    \centering
    \begin{tabular}{@{} l cccc @{}}
    \toprule
     Feature & \lang{bzd} & \lang{gn} & \lang{quy} & \lang{shp} \\
     \midrule
     Number of examples (Parallel) & 7,508 & 26,032 & 121,064 & 14,592 \\
     Number of examples (Wiki) & - & 4721 & 22610 & - \\
     \midrule
     Number of tokens - MLM-T   & 123,992 & 1,104,645 & 3,912,582 & 179,451 \\
     Number of tokens - TLM   & 194,798 & 2,006,996 & 6,697,771 & 328,427 \\
     Number of tokens - Wiki & - & 1,460,240 & 2,023,297 & - \\
    \midrule
    Number of dev examples & 50 & 48 & 45 & 46   \\
    Number of test examples & 50 & 50 & 50 & 50 \\
    \bottomrule
    \end{tabular}
    \captionof{table}{Features of the data used for our experiments.}
    \label{tab:parallel_data_details}
\end{table}

\section{Supplementary Results}

\begin{table}[h]
\centering
\setlength{\tabcolsep}{5pt}
    \begin{tabular}{@{} l l rrrr @{}}
    \toprule
    Model & Method & \lang{bzd} & \lang{gn} & \lang{quy} & \lang{shp} \\
    \midrule
    AWESoME  & BL  & 65.38 & 58.51 & 63.98 & 62.80  \\
    +mBERT         & +MLM-T & 64.26 & 43.29 & 39.10 & 66.44  \\
                   & +MLM-ST & 65.43 & 43.46 & 37.20 & 65.63  \\
                   & +TLM   & 54.25 & 34.62 & 30.38 & 62.10 \\
    \addlinespace 
    AWESoME  & BL  & 76.29 & 71.85 & 71.53 & 73.96 \\
    +XLM-R         & +MLM-T & 72.73 & 57.50 & 43.30 & 69.25 \\
                   & +MLM-ST&73.08 & 60.28 & 44.88 & 70.48\\
                   & +TLM   & 71.88 & 49.76 & 36.11 & 69.23\\
    \addlinespace
    FastAlign  & Union &47.39 & 39.78 & 58.37 & 57.91\\
    Giza++ & Union & 51.03 & 62.07 & 47.18 & 64.98 \\
    \bottomrule
    \end{tabular}
    \captionof{table}{Development AER for each language and method.}
\label{tab:main_dev_results}

\end{table}

\begin{table}[h]
    \centering
    \begin{tabular}{@{} ll rrrr @{}}
    \toprule
     Method & Heuristic & \lang{bzd} & \lang{gn} & \lang{quy} & \lang{shp} \\
     \midrule
 FastAlign & grow-diagonal-final & 54.56 & 49.64 & 60.51 & 56.11 \\
&grow-diagonal & 55.36 & 50.41 & 63.81 & 56.87 \\
&intersection & 57.11 & 52.89 & 66.92 & 61.67 \\
&union & 51.40 & 43.52 & 54.06 & 54.67 \\
&reverse & 52.21 & 51.51 & 61.27 & 58.41 \\

  \addlinespace
  Giza++ & grow-diagonal-final & 55.51 & 53.38 & 75.29 & 62.72 \\
 &grow-diagonal & 59.33 & 58.41 & 80.06 & 69.53 \\
 &intersection & 63.71 & 64.95 & 82.55 & 77.41 \\
 &union & 55.61 & 49.92 & 66.01 & 60.84 \\
 &reverse & 56.43 & 62.20 & 76.05 & 72.39 \\
    \bottomrule
    \end{tabular}
    \captionof{table}{AER results on the test set for various growing heuristics. }
    \label{tab:growing_heuristics}
\end{table}

\begin{table}[th]
\centering
\small
\setlength{\tabcolsep}{5pt}
    \begin{tabular}{@{} l l rrr rrr rrr rrr @{}}
    \toprule
    & & \multicolumn{3}{c}{{\lang{bzd}}}  & \multicolumn{3}{c}{{\lang{gn}}}  & \multicolumn{3}{c}{{\lang{quy}}}  & \multicolumn{3}{c}{{\lang{shp}}}  \\
    \cmidrule(lr){3-5} \cmidrule(lr){6-8} \cmidrule(lr){9-11} \cmidrule(lr){12-14}
    Model & Method &P&R&F&P&R&F&P&R&F&P&R&F \\
    \midrule
    AWESoME & BL & 41.8 & 23.4 & 30.0 & 50.6 & 29.0 & 36.9 & 49.4 & 24.7 & 33.0 & 64.0 & 28.7 & 39.6 \\ 
    (mBERT) & +MLM-T & 42.7 & 24.4 & 31.1 & 68.6 & 39.7 & 50.3 & 69.1 & 43.5 & 53.4 & 66.0 & 30.6 & 41.8 \\ 
    & +MLM-ST & 42.9 & 22.3 & 29.4 & 67.2 & 39.5 & 49.8 & 73.8 & 47.1 & 57.5 & 67.6 & 29.8 & 41.3 \\ 
    & +TLM & 62.1 & 31.2 & 41.6 & 76.3 & 45.4 & 56.9 & 79.0 & 52.5 & 63.0 & 79.4 & 34.0 & 47.7 \\ 
    \addlinespace
    AWESoME & BL & 48.0 & 12.5 & 19.9 & 48.4 & 18.6 & 26.9 & 49.7 & 16.5 & 24.8 & 64.5 & 20.2 & 30.8 \\ 
    (XLM-R) & +MLM-T & 38.9 & 16.4 & 23.1 & 63.0 & 23.8 & 34.6 & 70.2 & 34.6 & 46.4 & 57.2 & 25.0 & 34.8 \\ 
    & +MLM-ST& 40.8 & 15.5 & 22.5 & 65.7 & 24.3 & 35.4 & 74.8 & 34.4 & 47.1 & 56.5 & 23.7 & 33.4 \\ 
    & +TLM & 50.0 & 16.8 & 25.1 & 76.9 & 28.1 & 41.2 & 83.2 & 43.1 & 56.8 & 77.0 & 23.9 & 36.5 \\ 
    \addlinespace
    FastAlign & Union &46.4 & 51.0 & 48.6 & 55.4 & 57.6 & 56.5 & 44.3 & 47.7 & 45.9 & 48.0 & 43.0 & 45.3 \\
    Giza++ & Union &39.9 & 49.8 & 44.3 & 48.3 & 52.0 & 50.1 & 32.0 & 36.3 & 34.0 & 37.2 & 41.4 & 39.2 \\
    \midrule
     mBERT & +MLM-WT & -  & - & -  & 76.3 & 49.4 & 60.0 & 70.8 & 43.6 & 54.0 & -  & - & -    \\ 
     XLM-R & +MLM-WT& -  & - & -  & 66.4 & 31.5 & 42.7 & 75.0 & 38.8 & 51.2& -  & - & -   \\
    \bottomrule
    \end{tabular}
    \caption{Precision, recall, and F-measure for main test set results. All metrics are on a 0--100 scale (larger is better).}
\label{tab:prf_test}
\end{table}

\begin{table}[h]
    \centering
    \begin{tabular}{@{} r rrrr @{}}
    \toprule

        Num.\ Examples & +TLM & +MLM-WT & +MLM-ST & FastAlign \\ 
             \midrule
    50 & 67.58 & 66.89 & 67.53 & 67.26 \\ 
    100 & 67.74 & 63.87 & 65.79 & 66.97 \\ 
    200 & 68.42 & 65.28 & 65.75 & 66.91 \\ 
    400 & 65.61 & 66.43 & 65.31 & 66.80 \\ 
    800 & 63.43 & 62.84 & 63.76 & 66.26 \\ 
    1600 & 61.81 & 59.82 & 63.34 & 65.91 \\ 
    3200 & 56.93 & 57.41 & 62.99 & 64.75 \\ 
    6400 & 50.59 & 52.24 & 61.83 & 64.84 \\ 
    12800 & 43.98 & 52.14 & 61.18 & 63.06 \\ 
    25600 & 39.87 & 48.69 & 56.80 & 59.72 \\ 

    \bottomrule
    \end{tabular}
    \captionof{table}{AER for each method and subset used in the Subset Analysis.}
    \label{tab:subset_analysis_results}
\end{table} 

\begin{table}[h]
    \centering
    \begin{tabular}{@{} rr rr rr @{}}
    \toprule
            \multicolumn{2}{c}{{+MLM-WT}} & \multicolumn{2}{c}{{+TLM}} & \multicolumn{2}{c}{{FastAlign}} \\
            \cmidrule(lr){1-2} \cmidrule(lr){3-4} \cmidrule(lr){5-6}
            Avg.\ Char & AER & Avg.\ Char & AER & Avg.\ Char & AER \\
                         \midrule
            13.20 & 64.13 & 14.31 & 68.97 & 14.31 & 65.99 \\
30.29 & 63.13 & 31.20 & 61.61 & 31.20 & 64.89 \\
41.49 & 63.39 & 42.51 & 55.95 & 42.51 & 63.89 \\
50.19 & 62.19 & 51.45 & 54.73 & 51.45 & 64.19 \\
57.47 & 61.01 & 59.23 & 53.70 & 59.23 & 63.89 \\
64.20 & 59.07 & 66.44 & 49.12 & 66.44 & 62.15 \\
70.83 & 59.24 & 73.30 & 50.04 & 73.30 & 63.38 \\
77.12 & 57.06 & 80.09 & 48.12 & 80.09 & 63.56 \\
82.63 & 57.63 & 87.02 & 48.10 & 87.02 & 63.15 \\
89.31 & 55.77 & 94.31 & 47.63 & 94.31 & 62.96 \\
96.66 & 55.54 & 102.30 & 46.76 & 102.30 & 63.78 \\
104.76 & 54.40 & 111.48 & 46.07 & 111.48 & 62.99 \\
113.76 & 53.24 & 122.29 & 45.56 & 122.29 & 62.43 \\
124.33 & 51.07 & 135.93 & 45.62 & 135.93 & 61.87 \\
137.03 & 51.36 & 154.86 & 44.35 & 154.86 & 63.31 \\
152.70 & 50.43 & 195.18 & 42.55 & 195.18 & 62.23 \\
174.88 & 50.25 & - & - & - &  -\\
212.44 & 51.10 & - & - & - &  -\\
319.76 & 49.22 & - & - & - &  -\\

    \bottomrule
    \end{tabular}
    \captionof{table}{AER for each method and length group used in the Length Analysis. Average Chars represents the average number of characters per example, for each group.}
    \label{tab:length_analysis_results}
\end{table}

\begin{figure*}[h]
\centering
\includegraphics[width=\textwidth]{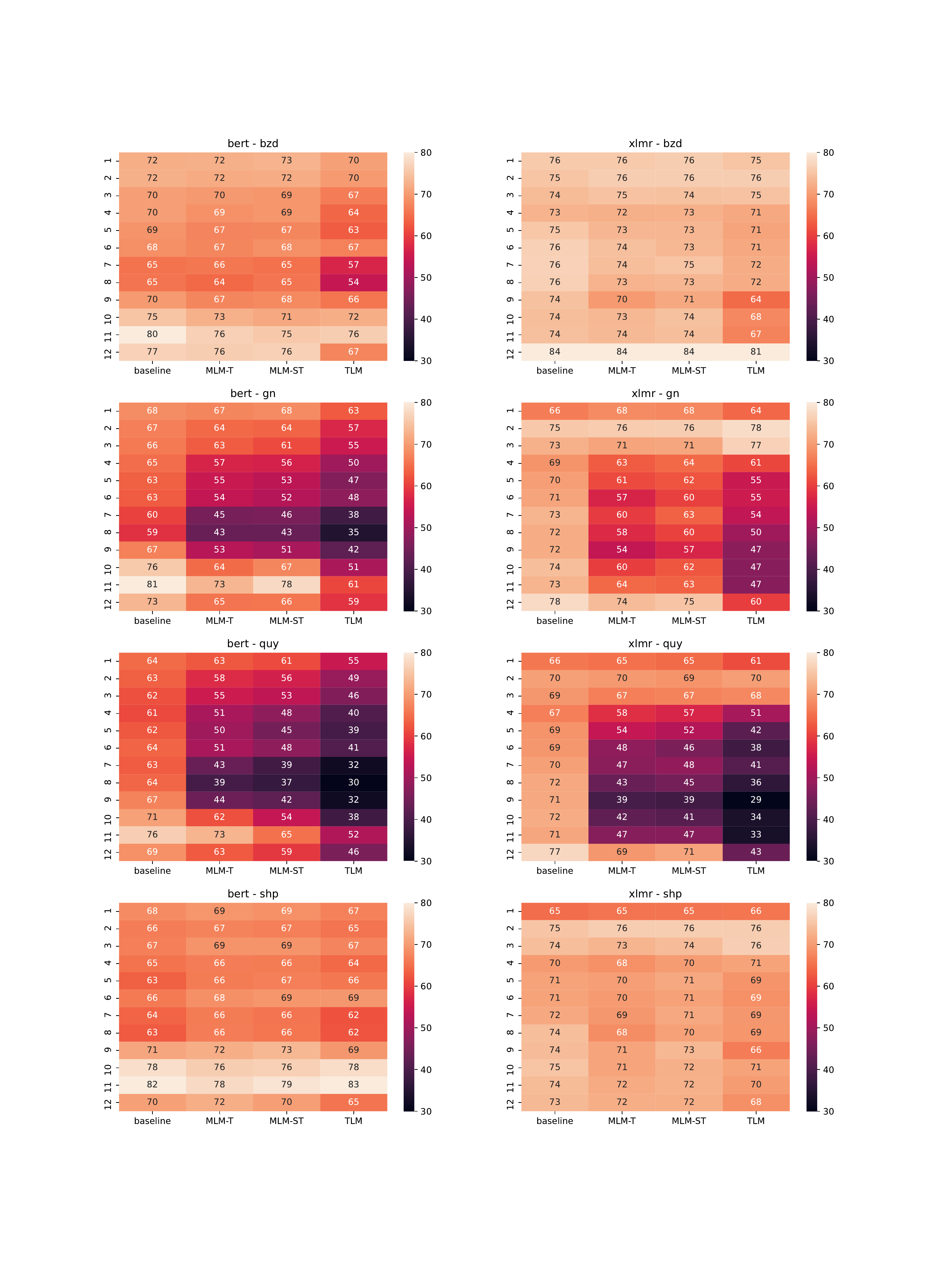}
\caption{AER using the development set, per layer, per language, for both mBERT and XLM-R.}
\label{fig:heatmap}
\end{figure*}

\end{document}